\documentclass[journal]{IEEEtran}
\pdfoutput = 1
\usepackage{amsmath,amsfonts}
\usepackage{algorithmic}
\usepackage{algorithm}
\usepackage{array}
\usepackage[caption=false,font=normalsize,labelfont=sf,textfont=sf]{subfig}
\usepackage{textcomp}
\usepackage{stfloats}
\usepackage{url}
\usepackage{verbatim}
\usepackage{cite}
\usepackage{textcomp}
\usepackage{paralist}
\usepackage[usenames, dvipsnames]{xcolor}
\usepackage{graphicx}
\DeclareGraphicsExtensions{.pdf,.jpg,.png,.eps}

\def\labfig#1{\label{fig:#1}}
\newcommand{\figref}[1]{Fig. \ref{fig:#1}}

\newcommand{\ctext}[1]{\raise0.2ex\hbox{\textcircled{\scriptsize{#1}}}}

\usepackage{enumitem}

\begin{document}

\title{Humanoid Robot RHP Friends: Seamless Combination of Autonomous and Teleoperated Tasks in a Nursing Context}

\author{Mehdi Benallegue, Guillaume Lorthioir, Antonin Dallard, Rafael Cisneros-Lim\'on, Iori Kumagai, Mitsuharu Morisawa, Hiroshi Kaminaga, Masaki Murooka, Antoine Andre, Pierre Gergondet, Kenji Kaneko, Guillaume Caron, Fumio Kanehiro, Abderrahmane Kheddar,~\IEEEmembership{Fellow,~IEEE,} Soh Yukizaki, Junichi Karasuyama, Junichi Murakami, Masayuki Kamon
\thanks{Manuscript received XXXX XX, 2024; revised XXXX XX, 2024.}%
\thanks{M. Benallegue, G. Lorthioir, A. Dallard, R. Cisneros-Lim\'on, I. Kumagai, M. Morisawa, H. Kaminaga, M. Murooka, A. Andre, P. Gergondet, K. Kaneko, G. Caron, F. Kanehiro and A. Kheddar are with CNRS-AIST Joint Robotics Laboratory, IRL3218, Tsukuba, Japan.}
\thanks{A. Dallard and A. Kheddar are also with the CNRS-University of Montpellier, LIRMM, UMR5506, Montpellier, France.}
\thanks{G. Caron is also with University of Picardie Jules Verne, MIS lab, Amiens, France.}
\thanks{S. Yukizaki, J. Karasuyama, J. Murakami and M. Kamon are with Kawasaki Heavy Industries, Tokyo, Japan.}
}

\markboth{Journal of \LaTeX\ Class Files,~Vol.~14, No.~8, August~2021}%
{Shell \MakeLowercase{\textit{et al.}}: A Sample Article Using IEEEtran.cls for IEEE Journals}

\IEEEpubid{\makebox[\columnwidth]{0000--0000/00\$00.00~\copyright~2021 IEEE\hfill} \hspace{\columnsep}\makebox[\columnwidth]{}}

\maketitle

\begin{abstract}
This paper describes RHP Friends, a social humanoid robot developed to enable assistive robotic deployments in human-coexisting environments. As a use-case application, we present its potential use in nursing by extending its capabilities to operate human devices and tools according to the task and by enabling remote assistance operations. 
To meet a wide variety of tasks and situations in environments designed by and for humans, we developed a system that seamlessly integrates the slim and lightweight robot and several technologies: locomanipulation, multi-contact motion, teleoperation, and object detection and tracking.
We demonstrated the system's usage in a nursing application. The robot efficiently performed the daily task of patient transfer and a non-routine task, represented by a request to operate a circuit breaker. This demonstration, held at the 2023 International Robot Exhibition (IREX), conducted three times a day over three days.
\end{abstract}


\section{Introduction}
In many countries, the shortage of workers due to declining birthrates and aging populations is becoming a serious societal, economic, and political concern. In Japan, the working-age population is projected to decrease by approximately 590,000 annually\footnote{``{Population Projections for Japan: 2021 to 2070 (With long-range Population
  Projections: 2071 to 2120)},'' \emph{National Institute of Population and
  Social Security Research}, 2023. \url{https://www.ipss.go.jp/pp-zenkoku/e/zenkoku_e2023/pp_zenkoku2023e.asp}}
. As the number of elderly people increases, the number of required caregivers is expected to grow by approximately 30,000 each year\footnote{``{The number of long-term care workers necessary to implement the plan for the
  8th Long-term Care Insurance Project (Japanese)},'' \emph{Ministry of Health
  and Welfare of Japan}, 2021. \url{https://www.mhlw.go.jp/stf/houdou/0000207323_00005.html}} and therefore, the labor shortage is expected to become even more severe. As labor costs soar, it is becoming increasingly difficult to compensate for the shortage by acquiring human resources from overseas. Therefore, robots are expected to help address the labor shortage~\cite{Vogel_RAM_2021}. In Japan, the Ministry of Economy, Trade and Industry (METI) and AMED are implementing several projects to support the development and introduction of robotic nursing care devices. Several priority fields have been established in these projects, including transfer support, mobility support, excretion support, and monitoring. By specializing in individual fields, robotic nursing care devices are being developed to be sold at low cost.

These efforts need to be further promoted in anticipation of the growing workforce shortage.

Indeed, providing support in eating, bathing, and toileting are three primary responsibilities of caregivers. Each of these activities takes place in a different location, so frail individuals have to move between them. If a person has difficulty standing up and moving by her/himself, s/he needs to use a wheelchair or other means to move around, which requires transferring from bed to wheelchair, from wheelchair to toilet, and so on. Transfer support by caregivers places a heavy burden on the backs~\cite{Choi_WJNR_2024}, resulting in back-pain and causing caregivers to leave their jobs. Hence, transfer support has been designated as one of the priority challenges. This research aims to achieve transfer support using an autonomous humanoid robot.

\begin{figure}[!t]
\centering
\includegraphics[width=0.75\columnwidth]{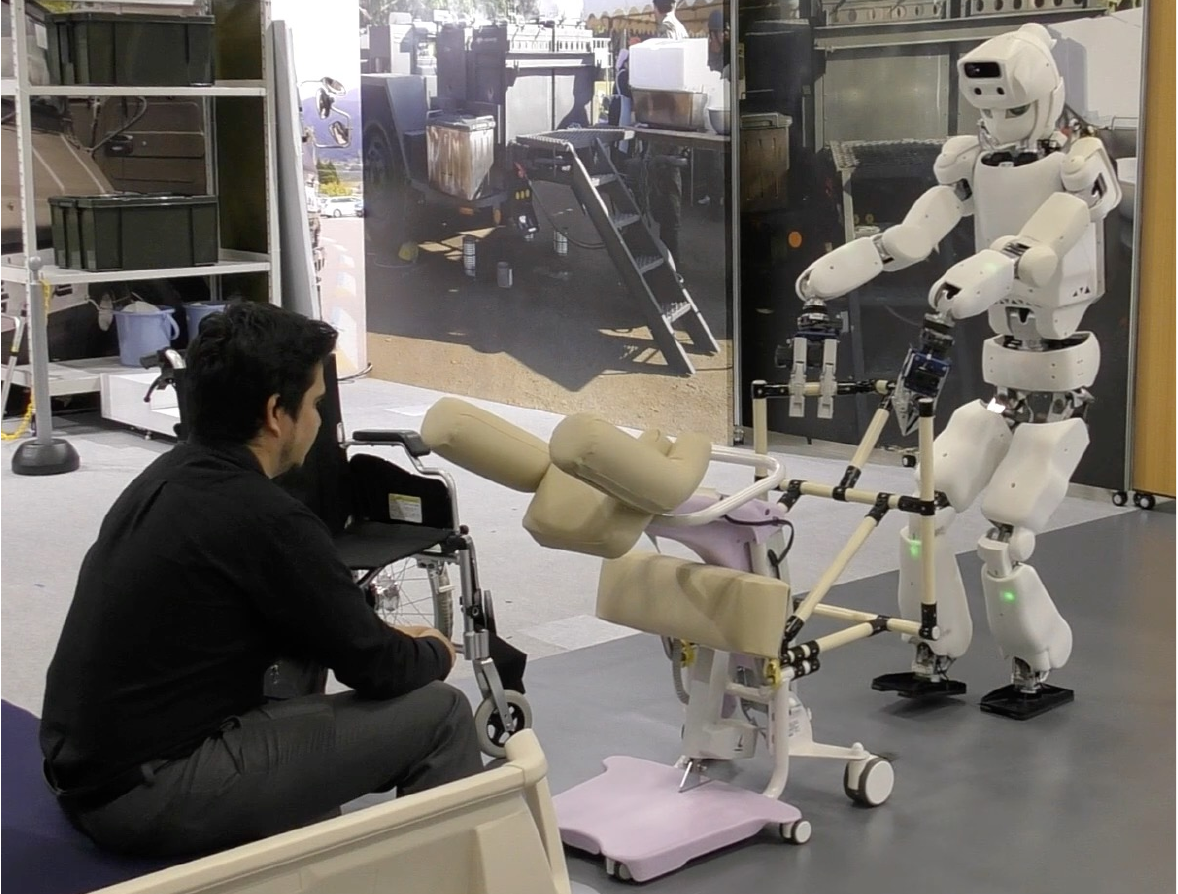}
\caption{RHP Friends, a social humanoid robot is locomanipulating Fuji's Hug L1-01, a transfer support device.}
\label{fig:friends_at_irex}
\end{figure}

Fuji's Hug series, Muscle's Sasuke, and Innofis' Muscle Suit are transfer support devices designed to reduce caregivers' physical burden, but they still require caregiver assistance. Autonomous robots like RIKEN's RIBA, RIBA II and ROBEAR\footnote{ \url{https://www.riken.jp/en/news_pubs/research_news/pr/2009/20090827/} \\ \url{https://www.riken.jp/en/news_pubs/research_news/pr/2011/20110802_2/} \\ \url{http://rtc.nagoya.riken.jp/ROBEAR/}  (In Japanese)} have also been developed for patient transfer. These robots, equipped with a human-like upper body on a mobile cart, tend to be large, heavy, and costly due to the need for high-power motors and a stable footprint to avoid tipping over. For instance, ROBEAR weighs 140~kg and measures 80~cm by 80~cm, making it too large to navigate standard doorways, which are typically 60~cm wide. Therefore, this study leverages one of the advantages of humanoid robots: the ability to expand their capabilities using tools. Instead of supporting the patient's weight with the humanoid robot itself, we will use transfer support devices developed and marketed for human caregivers to support transfers. This approach avoids increasing the size and weight of the humanoid robot while maintaining a footprint similar to that of a human.

The transfer support device we used is Fuji's Hug L1-01\footnote{\url{http://www.fuji.co.jp/items/hug/hugl1} (In Japanese)}(Fig.~\ref{fig:friends_at_irex}),  equipped with one actuator to support the patient when standing up during transfers and can be used for patients weighing up to $100$~kg. The wheels for transfer are passive and must be driven by a caregiver or a robot pushing the device.

In automobile automation and production sites, robots operate in well-structured environments with skilled employees working at a distance. In contrast, nursing facilities are less structured, with diverse patient needs, requiring closer human-robot interaction. Anticipating all scenarios and developing a robot that can autonomously handle every situation is challenging.
If a human can intervene remotely through teleoperation to assist with situations that the robot’s autonomy cannot handle, it reduces the need for a caregiver to be physically present for every incident. This is particularly beneficial during late-night hours when caregiver availability is limited. Remote intervention allows for scalable support, enabling operators with different skill sets to be called upon as needed, depending on the complexity of the situation. For example, a minor issue might only require voice communication or joystick navigation, while more technical or medical tasks may need the assistance of operators with specialized respective training. Following this trend, recent works in shared autonomy systems for caregiving show progress in safety and manipulability~\cite{Boguslavskii2023}.

This approach can also provide opportunities for people who are unable to participate in traditional caregiving roles due to family circumstances, geographical constraints, or physical limitations, allowing them to contribute remotely from their homes. By creating a diverse pool of operators that can be shared among multiple hospitals and caregiving institutions, this model promotes greater social inclusion and helps alleviate labor shortages~\cite{Tachi_ICARTESVE2019}. 

Although it is challenging to achieve ``versatility'', one of the advantages of humanoid robots is their potential to operate fully autonomously. This can be achieved by combining a humanoid robot with a physical structure similar to that of a human and with high-level cognitive functions possessed by humans. In this study, we construct an autonomous/remote hybrid system where routine nursing care tasks such as transfer support are performed by the autonomy embedded in the humanoid robot, while irregular tasks that are difficult to handle autonomously can be handled by remote support.

Currently, we assume that the switching between autonomous and remote modes is managed manually upon request, either by a monitoring agent using surveillance systems or through patient-initiated interactions, such as voice commands. This approach allows human oversight and intervention as needed. However, this does not exclude the possibility of integrating AI-based systems for automatic recognition of unusual situations in the future. Such systems, once reliable, could autonomously detect anomalies and seamlessly alert an available teleoperator to perform the switching.

\subsection{Contributions}

This work presents several key contributions to advancing humanoid robotics in nursing care through the integration and validation of existing technologies in a unified system:

\begin{itemize}[leftmargin=*,labelsep=0.5em]
    \item {Seamless Integration of Autonomous and Teleoperated Functionalities:} We integrate autonomous locomanipulation, multi-contact planning, control, and teleoperation capabilities into a cohesive framework, enabling the robot to perform both routine and complex tasks in a nursing care context. In addition, we present a system for switching in real-time between autonomous and teleoperated modes, enabling quick adaptation to dynamic situations and ensuring continuous operation and safety.

    \item {Real-World Validation:} The system was validated through live demonstrations at IREX 2023, showcasing its robustness under high strain real-world conditions in front of a large audience.

    \item {Practical Enhancements:} We implemented several technical improvements, including robust perception for object tracking, an intuitive teleoperation interface, and mechanisms for safely maneuvering loaded devices, addressing practical deployment challenges.

\end{itemize}

These contributions demonstrate the feasibility of deploying humanoid robots for nursing care by unifying existing technologies into a system capable of handling both routine and complex tasks in real-world environments.

\section{RHP Friends and its enhancements}
\label{sec:friends}
RHP Friends is a humanoid robot jointly developed by KHI, AIST, and CNRS. It is designed with the utmost safety in mind. Its slim, lightweight, and friendly appearance ensures it is not dangerous or frightening when working in close contact interaction with people.

\subsection{Humanoid robot RHP Friends}

\begin{figure*}[t]
 \centering
 \includegraphics[width=\textwidth]{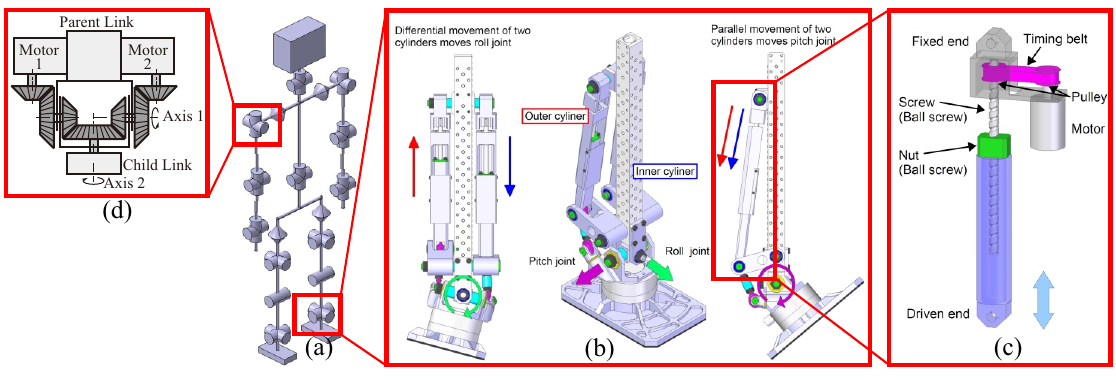}
 \caption{(a) Kinematic configuration of the humanoid robot RHP Friends, (b) a Closed Linkage Mechanism Unit, (c) a Linear Actuator Unit, and (d) a Differential System}
 \label{fig:RHPS-1}
\end{figure*}

RHP Friends\footnote{\url{https://kawasakirobotics.com/asia-oceania/blog/story_21/}} has 30 degrees of freedom (DoF) excluding the hands (see Fig.~\ref{fig:RHPS-1}(a)) with 6 in each leg, 7 in each arm, and 2 in the torso and neck. It is 1.68 m tall, weighs 54 kg, and operates using an onboard real-time control system on Linux. Fieldbus EtherCAT is used to communicate with motor controllers and sensors.

The two DoF joints in the legs, namely crotch and ankle joints, are driven with pair of linear actuators shown in Fig.~\ref{fig:RHPS-1}(c), which is also used in RHP2\cite{Kakiuchi_IROS2017}.
The advantage of this mechanism is that it has higher stiffness and strength against impulsive forces compared to conventional joint mechanisms using gear reducers, such as harmonic drive.
As shown in Fig.~\ref{fig:RHPS-1}(c), linear motion is produced by a ball screw.
A closed linkage mechanism is constructed using two linear actuator units, as shown in Fig.~\ref{fig:RHPS-1}(b).

The two DoF joints other than the crotch and ankle joints, shown in Fig.~\ref{fig:RHPS-1}(a), use a differential mechanism with bevel gears, as shown in Fig.~\ref{fig:RHPS-1}(d). The principle is as follows: When motors 1 and 2 turn in the same direction, the child link moves around axis 2. When motors 1 and 2 turn in opposite directions, the child link moves around axis 1. Arbitrary movement around axes 1 and 2 can be generated by combining the above movements.

The advantage of these mechanisms is that the torque generated at the joint becomes the sum of both motors scaled by speed ratio.

\subsection{Enhancements}
\label{sec:friendsEnhancements}

To this basic specification of RHP Friends, we additionally implemented an enhanced system to achieve object detection and tracking (see Sec.~\ref{sec:objdettra}) as well as visual feedback for teleoperation (see Sec.~\ref{sec:teleoperation}). 
The head was equipped with an RGBD camera capable of capturing wide-angle depth maps (Microsoft, Azure Kinect) and a stereo camera (Stereo Labs, ZED mini).
The RGBD camera is mainly used for object detection, and the stereo camera is mainly used for providing a 3D stereographic image to operator's head-mounted display.
A second computer system using Jetson Orin NX was added under the left arm to capture and stream data from these sensors to the network.
We also installed two commercial grippers: Sake Gripper (https:// sakerobotics.com/) on each wrist (four in total).

\begin{figure}[t]
 \centering
 \includegraphics[width = 0.9 \columnwidth]{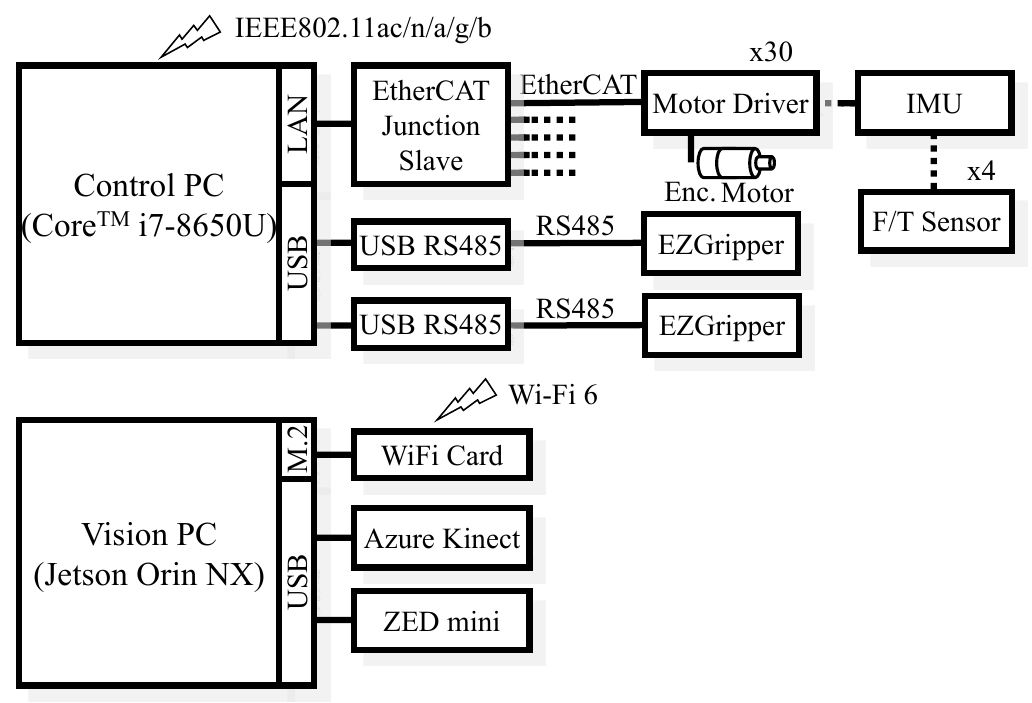}
 \caption{Electrical System of the RHP Friends humanoid robot.}
 \label{fig:electrical_system}
\end{figure}

Figure~\ref{fig:electrical_system} shows the electrical system of the enhanced RHP Friends.
The top part of Figure~\ref{fig:electrical_system} shows the basic RHP Friends computer system, and the bottom part shows the computer system added as part of the improvements.
Both computer systems communicate via WiFi.

\section{Software system overview}
\label{sec:softsys}
\begin{figure}[!t]
\centering
\includegraphics[width=\linewidth]{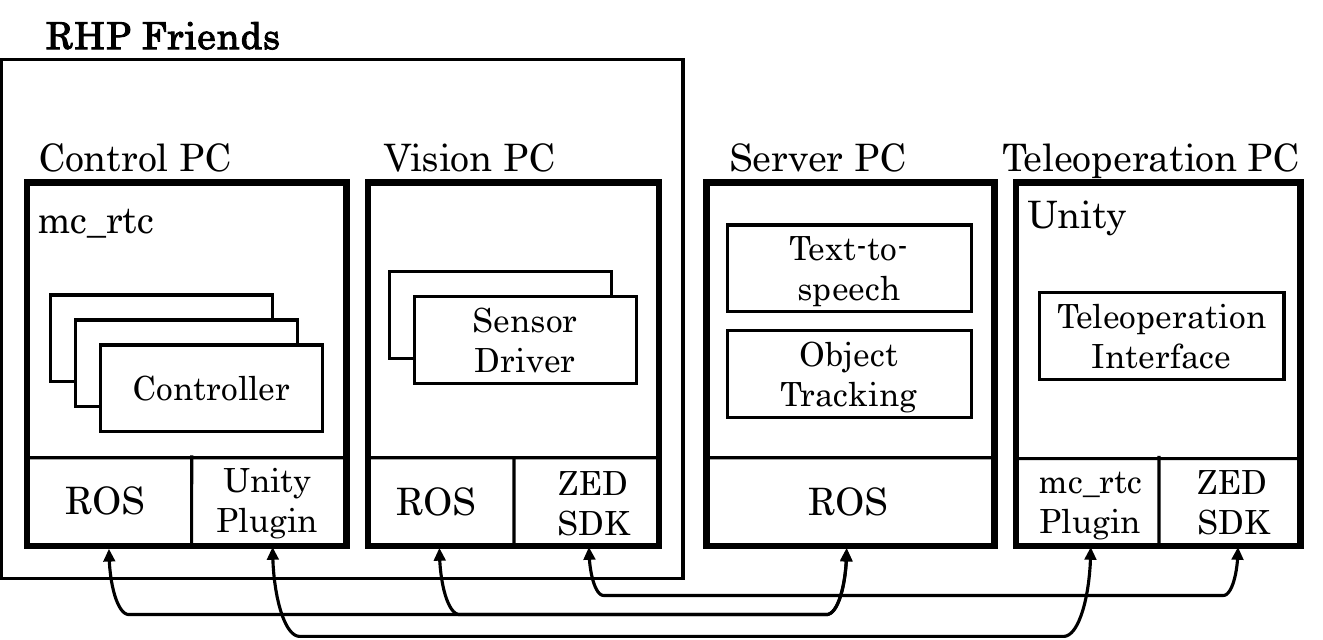}
\caption{The overall structure of the software architecture. The software components are distributed across four computers and communicate using ROS.}
\label{system_overview}
\end{figure}

Figure~\ref{system_overview} illustrates the software architecture distributed across four PCs using ROS for communication. The Control PC runs the whole-body motion controller, while the Vision PC processes data from the head-mounted Azure Kinect and ZED mini sensors. The Server PC handles object detection, tracking, and text-to-speech conversion using RGBD data from Azure Kinect (see Sec.~\ref{sec:objdettra}). The Teleoperation PC tracks the operator’s posture and displays ZED mini video streams with an overlaid user interface on the HMD (see Sec.~\ref{sec:teleoperation}).

The whole-body controller on the Control PC is implemented using mc\_rtc\footnote{\url{https://jrl-umi3218.github.io/mc_rtc/}}, a robot application framework developed since 2015. It features basic libraries for efficient real-time control and provides a unified interface for writing new controllers. With built-in statechart functionality configurable via text files, users can define complex controllers as tasks and constraints in a Quadratic Program (QP). mc\_rtc is independent of specific simulation platforms or middleware, offering a flexible simulation/control interface. This enables support for various robots with minimal programming and seamless use of the same controller for both simulation and testing. A list of supported robots is available on the framework's website\footnote{\url{https://jrl-umi3218.github.io/mc_rtc/robots.html}}.

The framework also offers a suite of tools to streamline the development of robotic applications. These include extensive logging capabilities and log visualization tools, a simple yet powerful API for building a user interface to visualize and modify the controller's internal state in real-time, and numerous extension points for various functional enhancements.

Using these features, mc\_rtc provides a development environment that enables newcomers such as interns, doctoral students, and post-doctoral fellows to start their research quickly.
mc\_rtc is open-source and used by many teams around the world. It makes it possible to efficiently implement complex robotic applications for large-scale projects such as COMANOID\cite{kheddar:ram:2019} and ANA Avatar XPRIZE\cite{cisneros:ijsr:2024}.
In addition to being able to switch controller configurations on a state-by-state basis using the controller's built-in statechart functionality, the controller itself can be seamlessly switched online to another controller.
This is because the states (of each controller) are just a collection of active tasks and constraints using a common solver.
This capability facilitates parallel development by using different controllers for functions that are used exclusively or by reusing controllers developed in the past. In fact, the technologies described in Sections~\ref{sec:locomani}, \ref{sec:multicontact}, and \ref{sec:teleoperation} are implemented as separate controllers.

\section{Locomanipulation}
\label{sec:locomani}
This section describes a method for generating and stabilizing the motion RHP Friends requires to maneuver a transfer support device (TSD) to a particular destination.
This motion can be described as loco-manipulation, a form of manipulation involving locomotion.

Typically, the TSD must be positioned in front of a patient so that they can be lifted with the device, carried to the location of a wheelchair, and transferred to it.
In such a scenario, the end-effector force required by the RHP Friends humanoid robot to maneuver the TSD differs significantly depending on whether the device is supporting the patient or not.
Although the TSD is equipped with freely rotating casters that enable it to move in any direction, the load becomes significant when the device is supporting the patient. As such, the center of rotation cannot be freely specified due to the constraints of the grasping force exerted by the robot's hands.

The manipulation force of the TSD depends on the weight of the patient and friction between the wheel of the TSD and the floor. We measured the manipulation force in advance from various people, and provide it as a desired wrench to a admittance control on the hands. The difference between the desired and measured wrench modifies the hand position by admittance control until the manipulation force and reaction force are balanced. When a person is on the TSD, most of the cameras's view on the robot is blocked.
To address these uncertainties, the robot adjusts its position using object recognition, to ensure the TSD reaches its destination. Similar past research has been done on such as transporting wheelchairs with human, but transport to a specific location with visual information on the robot has not been considered ~\cite{Nozawa_Humanoids2011}.

\subsection{Perception-aided object maneuvering}

\subsubsection{Limitation of grasping posture}

Locomanipulation is subject to many constraints compared to free walking. First, we face the problem of where on the TSD the robot should grasp and what posture is appropriate. We looked for a suitable grasping posture that would allow maneuvering (or transporting) the TSD without changing the grasping location. Although it is possible to change the grasping before reaching the joint limits in principle, we avoid such an action because it would increase the transportation time. So, both hands are constrained to the device while the TSD is being maneuvered. Due to the gap between the torso sway and the position profile of the TSD, the grasping posture must be determined to ensure the joint range of the arm has sufficient margin during locomanipulation.

After supporting the patient, the TSD is maneuvered to the wheelchair through a sequence of steering motions (e.g., backward, turning, and straight). In addition to pushing and pulling the TSD with both hands, the required manipulation must generate tangential forces while turning. These manipulation forces are even more significant if the TSD supports the patient, and they can be generated by moving the center of mass (CoM) in the direction of the desired force. This strategy requires an even greater margin in the joint range of the arms. Based on these constraints, appropriate robot posture and grasping position/orientation were determined by trial and error through dynamic simulation\footnote{We use the simulation provided by Choreonoid (\url{https://choreonoid.org/en/}).}.

The TSD supports the patient by lifting them while they lean forward onto the device's handrail. Although the robot could grasp the original handrail of the TSD, an extra handrail is attached for safety reasons to prevent possible contacts between the arm of the robot and the user's face when the robot releases its hands from the handrail.

\subsubsection{Footstep generation from desired waypoints}

\begin{figure}[!t]
\centering
\includegraphics[width=1.0\columnwidth]{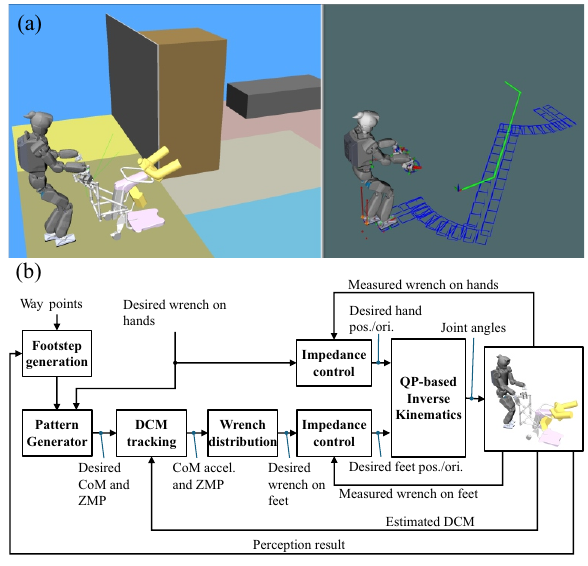}
\caption{(a) Example of footstep generation. The left figure shows the maneuvering motion of the TSD in the dynamics simulator Choreonoid. The right figure shows a foot placement sequence generated from predefined waypoints. (b) Locomanipulation control system.}
\label{fig:transport_tad}
\end{figure}

To reduce the required manipulation force as much as possible, the robot needs to maneuver the TSD supporting (loaded with) the patient using straightforward and turning motions with respect to an approximate position of the projection of the loaded TSD's CoM to the ground. The TSD is regarded as a nonholonomic differential two-wheeled robot and is first given a waypoint to follow a straightforward direction, as shown in Fig.~\ref{fig:transport_tad}(a).
The TSD is maneuvered by alternately shifting between straight and turning motions. We set a transit time for each waypoint, and the desired path of the TSD is generated using a trapezoidal velocity profile interpolation between the waypoints so that the TSD can be moved at a constant speed as much as possible. This strategy determines the relative foot placement from the future time trajectory of the TSD. When the patient is on the TSD, the CoM is approximately at the center of the foot as a center of turning. The TSD motion is generated less acceleration in the forward/backward direction except at the start/stop, and less lateral acceleration when turning. These movements prevent the person from experiencing sudden or lateral acceleration, and becomes consequently comfortable.

\subsubsection{Position adjustment via perception}
\label{sec:positionadjustment}

In a typical scenario, the robot maneuvers the TSD from several meters away to the front of the patient sitting on the bed.
Once the TSD supports this patient, it is maneuvered in front of the wheelchair, and the patient is lowered.
When a patient gets on the TSD, a large portion of the robot's camera field-of-view of the robot is occluded by the patient.
This situation makes it difficult to measure the wheelchair's position.
Thus, the static bed is used as a reference to estimate the robot's pose in the environment using high-speed 3D model-based tracking in wide-angle depth maps (see Sec.~\ref{sec:objdettra}) captured by the color-depth camera mounted in the robot's head (Sec.~\ref{sec:friendsEnhancements}).
Thus, when the bed comes into the capture range of the camera, the robot's position or direction is corrected at a pre-designated way point.
This enables the robot to maneuver the TSD in front of the person sitting on the bed, which is permanently tracked.
Then, similar position and direction corrections are performed to take the patient to the wheelchair. For example, after moving backward to reach the necessary distance to the bed to maneuver the TSD and after rotating the TSD toward the wheelchair to define the target pose to reach to let the patient sit on it.

\subsection{Stabilization for loco-manipulation}

The fundamental algorithm proposed in~\cite{murooka:ral:2021} is applied to stabilize the TSD while maneuvering it.
The control system for locomanipulation is shown in Fig.~\ref{fig:transport_tad}(b).
Maneuvering the loaded TSD requires different manipulation forces at the beginning and during the motion.
Since the joint range is highly restricted while maneuvering the TSD, the external forces are compensated by a feed-forward approach, in which the previously designed manipulation force profile is reflected onto the reference Zero-Moment Point (ZMP) trajectory.
The CoM trajectory is generated by preview control using this ZMP reference.
Then, the CoM position is stabilized by controlling the reaction force on the feet through Divergent Component of Motion (DCM) tracking.

\section{Object detection/tracking}
\label{sec:objdettra}

Accurate 3D visual tracking is crucial for precise robot manipulation in dynamic environments like nursing facilities.

Object 3D visual tracking is implemented to estimate the position and the robot's orientation with respect to a reference object at the camera frame rate. The nursing context offers several possible objects to consider, ranging from the smallest, such as a chair, to larger ones, such as a shelf, a table, or a bed. In this work, the only static object in the environment during the transfer operation is a bed, which is used as the reference with respect to which the locations of the patient, the wheelchair, and the robot waypoints are defined.

\begin{figure}[t]
 \centering
 \includegraphics[width=0.9\linewidth]{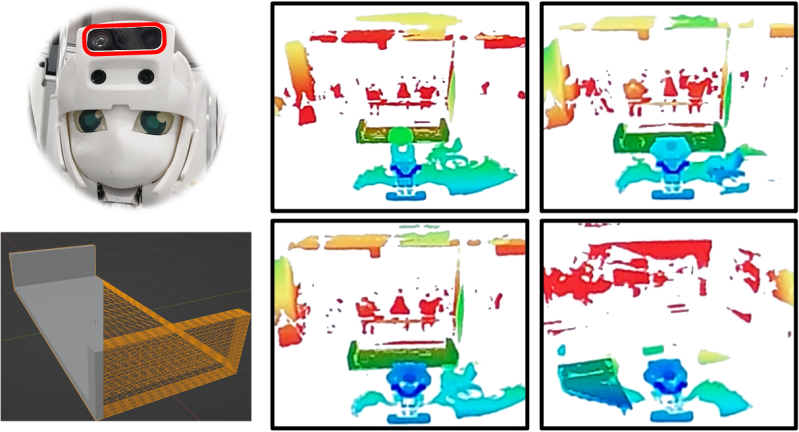}
 \caption{The first column shows the integration of the Azure Kinect camera in the robot's head (bordered with red) and the bed CAD model used. The two other columns show excerpts of dense bed 3D model-based tracking results in wide-angle depth images. They are shown as 3D point clouds with false colors from a third-person view. The color illustrates the distance from the camera (blue: close; red: far). The 3D model of the bed is shown in gray. The TSD is visible at the bottom of every view, which from left to right and top to bottom show tracking results when the robot reached: (i) the location where the bed is in the range of the depth camera; (ii) the patient on the bed; (iii) the backward location when to start rotating toward facing the wheelchair that is on the right side of the view; (iv) the location where the robot only has to go forward for making the patient seat on the wheelchair (note the bed is only partly visible on the bottom left side of the view despite the 120-degree field-of-view, although not preventing the tracking from succeeding.}
\label{fig:objdettra}
\end{figure}

Since a bed is a large object of about $2 \times 1$~m horizontal size with a bed foot and head of about $40$~cm height with a pedestal of the same height, it is necessary for the robot camera to feature a large field-of-view in order to see the bed even close to it or when the head is not directed to it, such as when the robot maneuvers with the loaded TSD. Furthermore, beds, particularly the standard ones, have a uniform visual appearance, which is challenging to track accurately and robustly with grayscale or color cameras. This double challenge for the object being large and of uniform appearance makes it hard for most of the plethora of visual pose estimation methods to work since they deal with narrow field-of-view, even though it is clear that the depth modality of color-depth cameras helps~\cite{Thalhammer_TRO_2024}. That is why we leverage our previous work on large industrial object locomanipulation with humanoid robots~\cite{Chappellet_TASE_2023}. This particular research led us to equip the humanoid head with a wide-angle depth camera (see Fig.~\ref{fig:objdettra}, top left) and develop fast, dense 3D model-based tracking in the depth images.

In practice,~\cite{Chappellet_TASE_2023} showed that even with a Computer-Aided Designed (CAD) 3D model approximating the actual object of interest, accurate tracking results can be obtained thanks to the explicit depth measured at scale by the depth camera. Hence, for tracking the bed, we made a simplified CAD model of the real bed with no detail (see Fig.~\ref{fig:objdettra}, bottom left). There are two practical advantages of working successfully with such coarse object models: (i) such CAD model is made in a matter of minutes, and (ii) it has the potential to be used with objects of slightly different details, e.g., a bed with different accessories in the bed head or foot if it is roughly of the same whole size.

Thus, the goal of the tracker is to estimate the ${^c{\bf M}}_o \in \mathrm{SE}(3)$ frame change from the object frame $\mathcal{F}_o$ to the camera frame $\mathcal{F}_c$ to align the best the CAD model of the object with each successively captured depth map. The optimal ${^c\hat{\bf M}}_o$ obtained for one captured depth map initializes the ${^c{\bf M}}_o$ for the next captured depth map. Within a multi-resolution approach to deal with coarse initial ${^c{\bf M}}_o(0)$, for example due to large inter-frame motion, \cite{Chappellet_TASE_2023} computes iteratively in a dense projective iterative closest point approach in three main stages. First, the 3D model is projected into the 2D distorted image plane of the wide-angle depth camera using the camera's intrinsic parameters. Next, the closest corresponding points are identified within a defined search range, considering both Euclidean distance and angular differences between surface normals. Finally, the frame transformation ${^c{\bf M}}_o$ is optimized by minimizing the 3D point-to-plane distance between the matched points, utilizing surface normal information 

Finally, the kinematic chain is leveraged to transform the robot coordinate system to the object one in order to compute the distance and the relative orientation between the robot and the bed for further robot position adjustment (Sec.~\ref{sec:positionadjustment}). 

\section{Multi-contact planning and control}
\label{sec:multicontact}
\begin{figure}[t!]
 \centering
 \includegraphics[clip,width=\columnwidth]{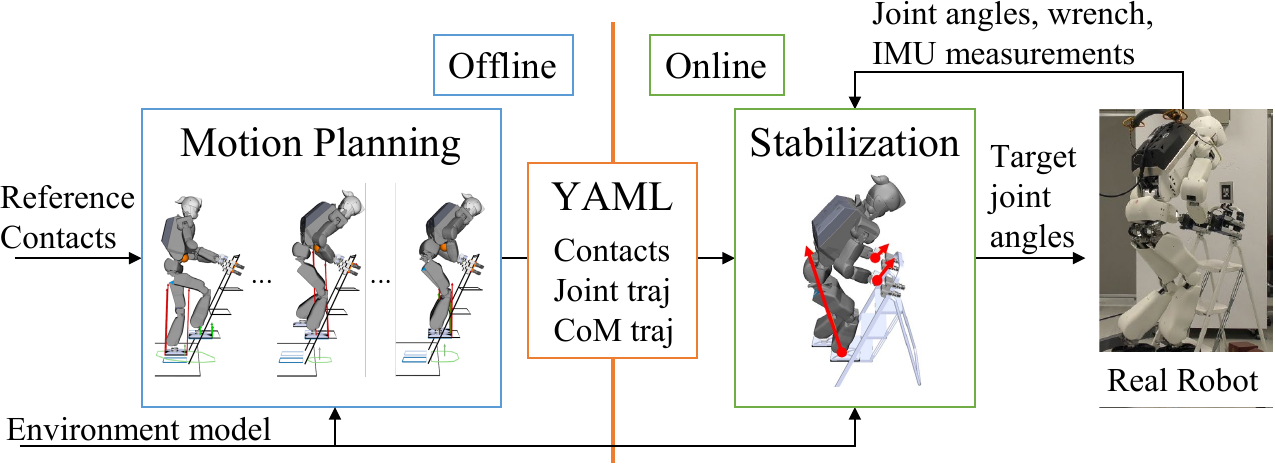}
 \caption{Overview of the multi-contact motion planning and control framework.}
 \labfig{multicontact_system_overview}
\end{figure}

In nursing works, a humanoid robot may be required to perform multi-contact tasks, 
where it simultaneously uses its legs and arms, leveraging its human-like body structure.
Multi-contact tasks for a humanoid robot have been developed focusing on its autonomy\cite{Kumagai2023},
but expert instructions are necessary in caregiving situations.
Recently, a teleoperation framework for multi-contact scenarios\cite{McCrory2023} was also proposed.
However, it is still difficult for caregivers who are non-experts in robotics to perform
multi-contact tasks considering the kinematics and statics of a robot.
Therefore, we developed a framework to achieve multi-contact motion planning and control
for a humanoid robot, which is shown in~\figref{multicontact_system_overview},
and integrated it with the intuitive teleoperation framework.
We assume that the environment model is available and we manually plan the target contact sequence in advance.

\subsection{Offline multi-contact motion planning}
First, we generate offline quasi-static whole-body motion of the target contact sequence. 
The latter is given by the optimization-based multi-contact motion planner described in~\cite{Kumagai2021}.
We assume that the robot moves one limb at each contact transition. 
Each transition motion is a result of inverse kinematics (IK) computed as an optimization problem for discretized frames of the transition motion.
When a humanoid robot performs multi-contact tasks, bilateral contact forces can expand the support area for its center of mass (CoM).
This enables the robot to move CoM away from the step-ladder, which helps it to avoid collisions between its legs and steps.
Our motion planner approximates bilateral contacts as pairs of unilateral surface contacts. This approach allows us to apply static equilibrium evaluation, commonly used for unilateral contacts.
After generating the quasi-static multi-contact motion, we preserve its contacts, joint trajectory, and CoM trajectory to use as references for the real-time multi-contact controller.

\subsection{Online multi-contact stabilization control}
When performing multi-contact tasks in the real-world, a humanoid robot dynamically maintains its balance while interacting with the environment with its arms and legs. 
To achieve this, we devised a multi-contact controller~\cite{Murooka2022} consisting of online centroidal trajectory generation and centroidal stabilization control. The centroidal trajectory generation is formulated as preview control~\cite{Katayama1985} instead of model predictive control (MPC), which can significantly reduce the computation time.
Although preview control does not explicitly consider equality and inequality constraints, the resulting trajectory is feasible because the reference CoM trajectory generated by the whole-body motion planner in~\cite{Kumagai2021} embeds feasibility constraints and wrench projection.
We then implemented the centroidal stabilization control to reduce errors between the desired and the actual CoM states and compensate for disturbances.
We compute the amount of wrench modification using a PD controller and distribute it among the contact wrenches considering contact constraints, which are achieved by damping control on each limb.
We manually tuned the control parameters for damping control and used the same one to both climb up and down the step-ladder.

\subsection{Assessment by climbing a step-ladder}
\begin{figure}[t!]
 \centering
 \includegraphics[clip,width=\columnwidth]{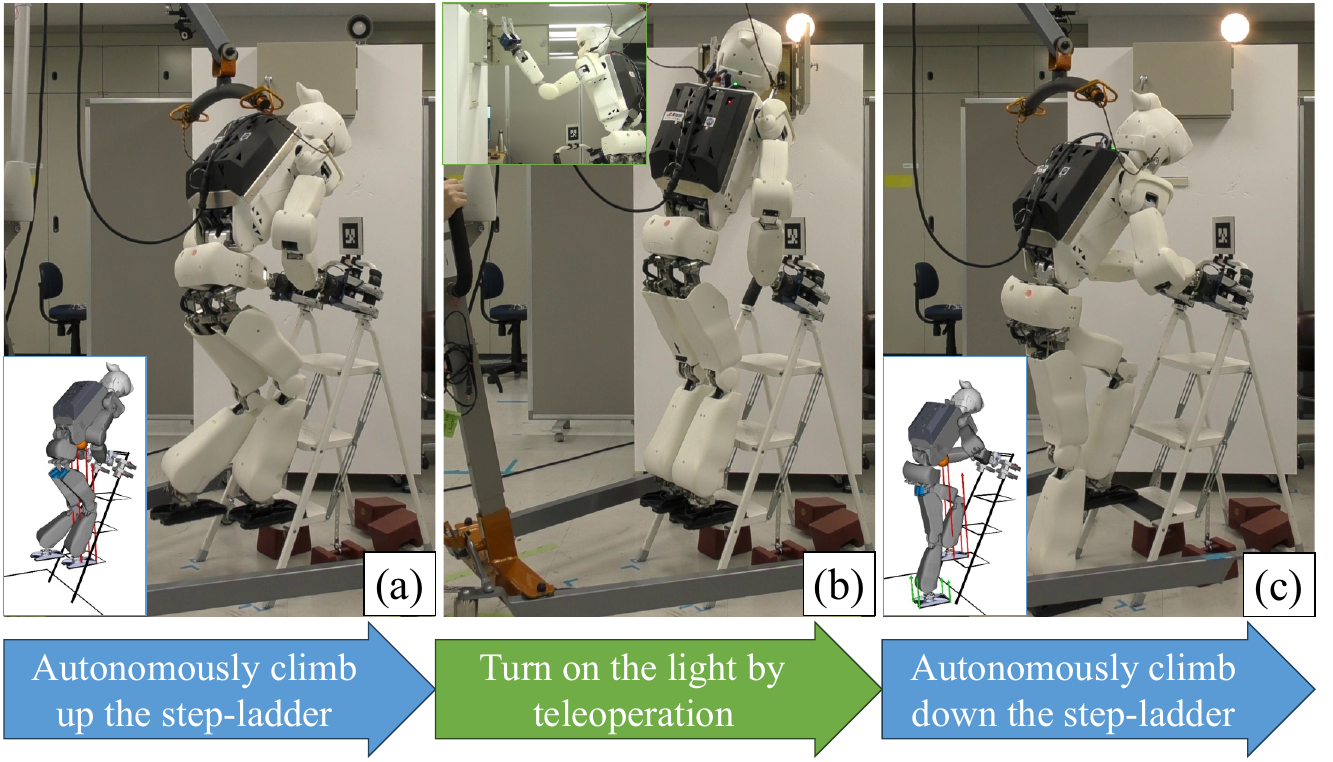}
 \caption{RHP Friends turned on the light using the step-ladder.
 (a) RHP Friends autonomously climbed up the step-ladder using the proposed multi-contact locomotion framework.
 The lower left figure is the planning result. (b) RHP Friends turned on a switch in the breaker box by teleoperation.
 The upper left figure magnifies the teleoperated arm. (c) RHP Friends autonomously climbed down the step-ladder
 using the proposed multi-contact locomotion framework. The lower left figure is the planning result.}
 \labfig{stepladder_experiment}
\end{figure}

We evaluated the proposed framework in an experiment where RHP Friends climbed up and down a step-ladder to turn on the light.
The results of this experiment are shown in~\figref{stepladder_experiment}.
The proposed framework successfully generated quasi-static reference motion that utilized bilateral contact forces while grasping handrails.
The reference motion is given to the stabilization controller that dynamically balances the robot while climbing up and down the step-ladder in the real world.
Note that the robot was teleoperated on the step-ladder when it turned on a switch in the breaker box; that is, once the robot had climbed up the step-ladder, the controller was switched online to the one for teleoperation (see Sec.~\ref{sec:teleoperation}), and then switched back to the multi-contact controller for climbing down.
This capability of switching controllers was explained in Sec.~\ref{sec:softsys}.
From the above results, we conclude that our proposed framework can contribute to expanding a humanoid robot's capabilities in nursing facilities.

Building on this approach, our recent work from leverages tactile sensing under the robot's feet to further enhance stability and adaptability during stair climbing, providing a complementary method to improve robustness in dynamic environments~\cite{tako:humanoids:2024}.

\section{Teleoperation}
\label{sec:teleoperation}
\begin{figure}[t]
	\centering
	\includegraphics[width = 0.80\columnwidth]{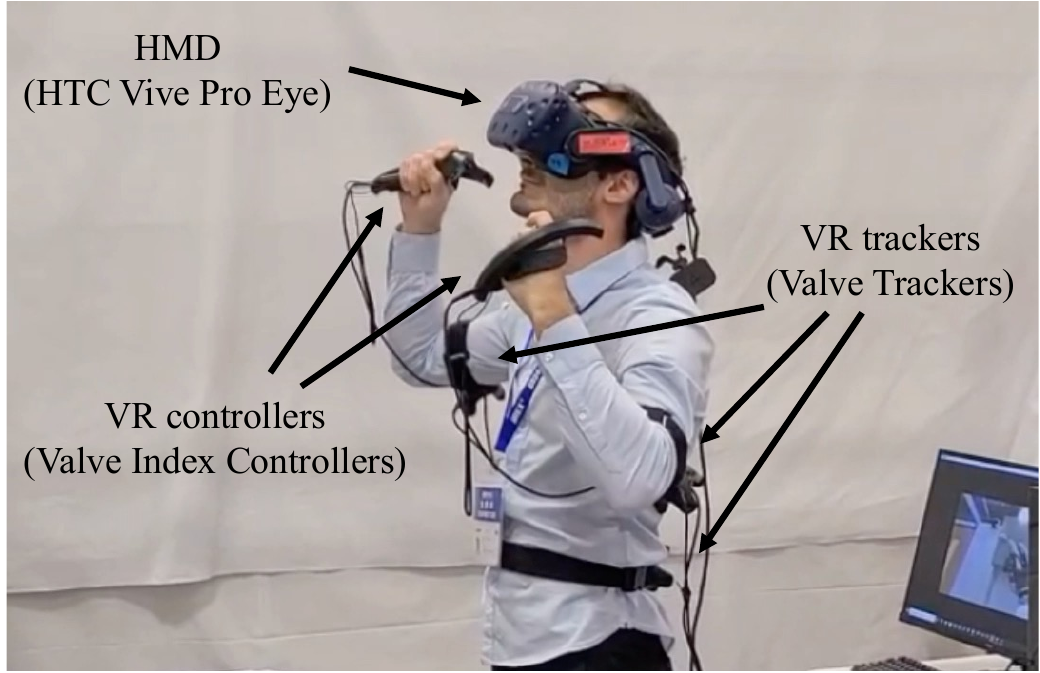}
	\caption{The operator system consists of a Head Mounted Display (HMD), a motion tracking system, and hand-held controllers.}
	\label{fig:operator}
\end{figure}

This section describes the teleoperation system that is used whenever a remote intervention is needed, allowing to turn the humanoid robot into a physical avatar that can be naturally controlled due to the morphological resemblance with the operator~\cite{Darvish_TRO2023}.

In healthcare facilities, a teleoperation system must be designed with the operator's convenience in mind to ensure practical adoption. Indeed, we are exploring approaches to adapt the system to operators with varying expertise levels, including our contribution to the ANA Avatar XPRIZE where we demonstrate that foundational expertise can be developed within an hour~\cite{cisneros:ijsr:2024}.

\subsection{Main architecture} \label{secmain-teleop-archi}

The teleoperation system is based on the one developed for the ANA Avatar XPRIZE competition~\cite{cisneros:ijsr:2024}, but adapted for the RHP Friends humanoid robot.

The operator system consists of a Head Mounted Display (HMD), a motion tracking system, and hand-held controllers, as shown in Fig.~\ref{fig:operator}.

The HMD (VIVE Pro Eye\footnote{\url{https://vive.com/us/product/vive-pro-eye/overview/}}) displays real-time images from the robot’s stereo camera, allowing the operator to see through the robot’s perspective.

This system is designed with flexibility in mind because the operator is not expected to be constantly engaged with the system. For instance, in case of simple navigation tasks, the system can be controlled using a monitor and the joystick interface, while more delicate tasks that require fine manipulation and precision are managed using the HMD and motion tracking system. Switching between these modes does not require any reconfiguration except turning on the trackers.

To track the operator's limbs in teleoperation mode, we used a motion tracking system consisting of 3 individual trackers (VIVE Tracker\footnote{\url{https://vive.com/us/accessory/tracker3/}}), each providing a 6D pose estimation.
They have been installed on the operator's

\begin{inparaenum}[(a)]
	\item lower back (to provide a reference frame with respect to which the motion of other limbs is defined) and
	\item elbows (to track the arm configuration as accurately as possible).
\end{inparaenum}

This technology requires the installation of base stations around the operator.

As for the hands, we used hand-held controllers (Steam: Valve Index Controllers\footnote{\url{https://store.steampowered.com/app/1059550/Valve_Index_Controllers/}}), which, besides providing tracking for the hands, offer a series of triggers, buttons, a track button, and a thumbstick.
These elements allow the operator to command walking motion, activate the hand control, and interact with the operator interface, as explained later in Sec.~\ref{sub:operator_interface}.

The avatar software framework controls the robot's whole-body by using the commands given by the operator through the HMD, the motion tracking system, and the hand-held controllers.
Data is transmitted through a server-client interface embedded in the framework instead of ROS; it also provides feedback from the robot to the operator over the network.

The whole-body control is formulated as a QP managed by a finite state machine (FSM) that receives inputs/commands from the operator.
Each state triggers a unique behavior with a different set of tasks (e.g., end-effector pose, force control) and constraints (e.g., self-collision, joint limits), thus implementing a control scheme that realizes our teleoperation framework.
The QP approach for robot teleoperation manages multiple constraints, including joint kinematics, self-collision, and contact-related limits, ensuring that generated motions remain feasible under these constraints. 
The details of the controller can be found in \cite{cisneros:ijsr:2024} the main difference is the balance and locomotion control, which is implemented as summarized in Sec.~\ref{sub:locomotion_control}.

\subsection{Operator interface}
\label{sub:operator_interface}

The framework developed for ANA Avatar XPRIZE \cite{cisneros:ijsr:2024} was tailored to meet the requirements of the competition, being one of them the haptic feedback.
As such, it prompted us to consider haptic gloves (SenseGlove DK1); that is, a button-less interface that drove the necessity of considering a vocal interface.
The resulting system was still simpler than other approaches that constrained the motion of the operator with grounded exoskeleton-based stations~\cite{Hauser_SORO2024}.
However, it required further simplification when having in mind the feasibility for it to be adopted at healthcare facilities, where efficiency is prioritized over embodiment of human telepresence.
This decision led to the adoption of Valve Index controllers, which offer accessible buttons and are easy to wear, and rendered the cumbersome button-less interface not mandatory.

The operator can enable/disable control of the robot’s head using the $A$ button on either Valve controller, and the arms with the $B$ button on the respective controller. The gripper closes continuously based on trigger pressure. Locomotion is managed with the joysticks: the left controls translations (forward, backward, sidestepping), and the right controls rotations. Both can be used together for complex patterns like walking in an arc. The touchpad buttons free the robot if the arm is stuck in a reversed shoulder joint configuration, restoring the maximum reachable space.

The operator's user interface has also been simplified.
It displays only the viewpoint captured by the robot's ZED camera, along with small icons indicating whether the control of the arms and head is currently enabled or disabled.
The F/T sensors situated on the robot's wrists are used to deliver pseudo-haptic feedback.
When the grippers come into contact with an object, the corresponding controller initiates vibrations, thereby alerting the operator about the contact with the environment.

\subsection{Balance and locomotion control in teleoperation}
\label{sub:locomotion_control}

The core of the walking control scheme is governed by the extended dynamics of the linear inverted pendulum model (LIPM).
This model reflects the relationship between the center of mass (CoM) acceleration and its displacement from the zero moment point (ZMP), a fundamental aspect of maintaining balance during walking.

Our control scheme tracks the desired linear and angular velocities the operator sends to navigate in the environment.
To do so, it dynamically adjusts the ZMP reference position through admittance foot force control, thereby influencing the CoM acceleration to maintain balance.
Additionally, the real-time feedback mechanism incorporated into the control scheme allows for continuous adjustment of both the CoM position and the desired ZMP position based on the observed state of the robot. 

Three core components constitute this controller:
\begin{itemize}[leftmargin=*,labelsep=0.5em]
 \item \textbf{Model-rich closed-loop feedback:} This controller integrates state feedback, including CoM state and ZMP location, measured with F/T sensors, into a dynamical model. This enables quick reactions to disturbances detected by the force sensors before they affect kinematics. It predicts LIPM trajectories and respects balance criteria without external stabilizers.
 \item \textbf{Dynamic re-planning:} The controller continuously re-plans step locations and timings using a cascade of solvers, ensuring optimal velocity tracking under varying conditions.
 \item \textbf{Feet force control:} Admittance control tracks forces exerted by the feet, maintaining force tracking stability and preventing oscillations in the system dynamics.
\end{itemize}

Refer to~\cite{dallard:hal04147602} for details and extensive experimental evaluations.
This control scheme has significantly enhanced our teleoperation framework, demonstrating valuable efficiency in adapting to changes in reference velocities and navigating uneven terrain.

\section{Evaluation of the system and experimental results}
To test the performance of the overall system, we devised a demonstration involving two tasks:

\begin{inparaenum}[(a)]
	\item a typical daily routine task at a nursing facility that could be performed autonomously, and
	\item a non-routine task that would require the intervention of an expert that would take control of the robot through teleoperation.
\end{inparaenum}

The routine task involves the RHP Friends humanoid robot (Sec.~\ref{sec:friends}) bringing the TSD and using it to transfer a patient from a bed to a wheelchair.
To do this, it is necessary to use the color-depth camera to detect the bed and track it in real-time (see Sec.~\ref{sec:objdettra}) while moving around holding the TSD with the hands; that is while locomanipulating it, first without load and then with the patient on board (see Sec.~\ref{sec:locomani}).

The non-routine task is a situation in which the patient requests work that requires specialized knowledge.
In this case, the request involves resetting one specific circuit breaker, among others, inside a breaker box
(and turning on a light for verification).
This task can be achieved by an operator overriding the control of the robot to teleoperate it (see Sec.~\ref{sec:teleoperation}).
Furthermore, if the breaker box is not at a reachable height, it is necessary to consider the use of a step-ladder.
Climbing up and down a step-ladder while holding its handrails requires multi-contact planning and control (see Sec.~\ref{sec:multicontact}).

\begin{figure*}[t]
\centering
\includegraphics[width = \textwidth]{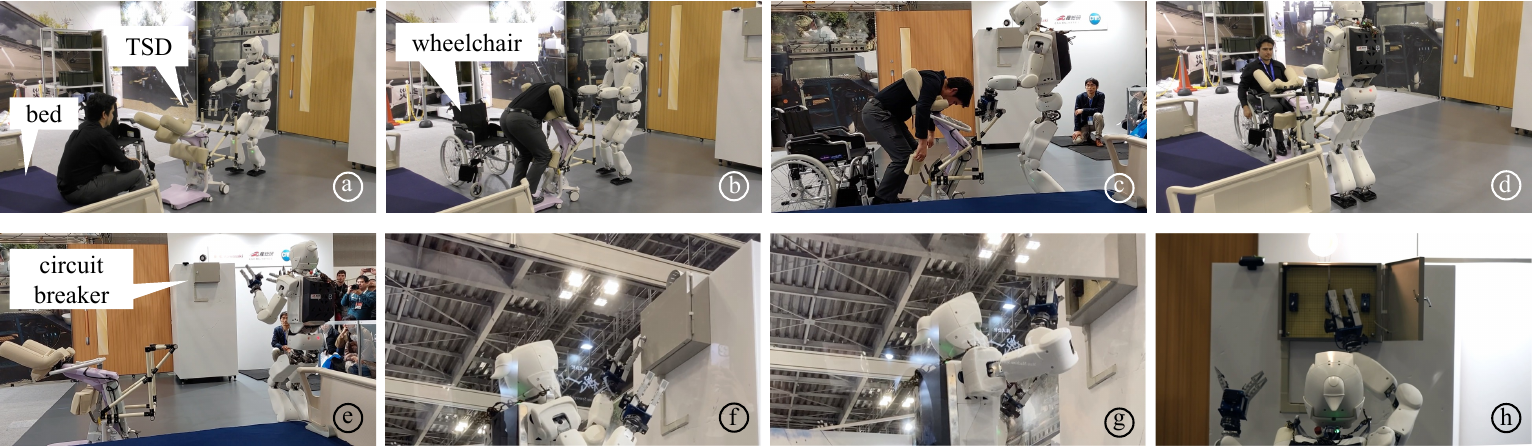}
\caption{Demonstration scenarios at IREX. \ctext{a} The bed is tracked, and the TSD is taken in front of the patient. \ctext{b} The TSD is operated wirelessly to support/lift the patient. \ctext{c} Turning the TSD with the patient on board toward the wheelchair. \ctext{d} The TSD is operated wirelessly, and the patient is lowered to sit in the wheelchair. Then, the patient asks to investigate a problem with the light. \ctext{e} The robot switches to teleoperation mode and moves to the front of the circuit breaker. \ctext{f} The robot opens the breaker box by turning the door handle. \ctext{g} The robot hooks the gripper on the lever and opens the breaker box door. \ctext{h} The robot operates the corresponding breaker switch to turn on the light.}
\label{fig:snapshots}
\end{figure*}

This scenario was demonstrated to the general public at the International Robot Exhibition (IREX\footnote{\url{https://irex.nikkan.co.jp/}}) 2023, held in Tokyo, Japan, for four days from November 29$^{th}$ to December 2$^{nd}$, 2023.
We performed the previously described demonstration at the Kawasaki Heavy Industries (KHI) booth every day of the exhibition and three times per day in front of the attendees.

At each demonstration, we showcased the routine task, as shown in Fig.~\ref{fig:snapshots}(a)-(e).
However, due to reliability concerns during the exhibition, we could only show the teleoperated non-routine task on two of the four days.
For the same reasons, we did not include the step-ladder part and we lowered the breaker box to a reachable height.
A snapshot of the teleoperation is shown in Fig.~\ref{fig:snapshots} (f)-(j).
However, as seen on the video of this demonstration\footnote{\url{https://www.youtube.com/watch?v=6hKvDK1HCRI}}, all other parts of the demonstration ran smoothly and reliably. 

Regarding the bed pose estimation (Sec.~\ref{sec:objdettra}), the first significant risk was  environmental perturbations for the depth camera: attendees, flash and infrared lights, and different materials around compared to the laboratory environment. During the preparation time, we noticed that the range of reliable depth measurement was slightly shorter than in the laboratory. Thus, we used a shorter distance to the bed (by 30 cm) for the first perception-based adjustment of the robot's position. Second, the bed shape posed a risk of failure due to its low-constrained shape on one axis (the horizontal axis of the header and the footer). However, the small part of the depth map on the mattress thickness (about $10$~cm height) is sufficient to avoid poor pose estimations. Third, since the tracker's efficient implementation requires a middle-class Graphics Processing Unit (GPU) to track an object in less than $5$~ms per captured depth map ($30$ Frames Per Second with the Azure Kinect using the Wide $120$ degrees Field-Of-View mode at $512 \times 512$~pixels), the Jetson Orin NX embedded in the robot was not powerful enough to run both the data capture and the tracking onboard. Thus, the data captured by the Azure Kinect was streamed through WiFi to the Server PC (a desktop PC with an Nvidia GeForce RTX 3060 Ti GPU). In the presence of up to 200 attendees per demonstration using their smartphones and other devices, possibly causing WiFi signal perturbations, thus data transmission issues caused object tracking failure. In the beginning, we indeed experienced WiFi communication issues. However, by judiciously placing an access point near the demonstration zone and adapting the WiFi channel, the data streaming performed sufficiently well to allow successful object tracking and the subsequent robot position adjustments. The interaction between the robot and the bed, as well as between the TSD and the bed during patient transport, and the precise final positioning of the TSD at the wheelchair, demonstrate that the object pose estimation was sufficiently accurate to complete these tasks safely~\footnote{The trapezoidal shape of the horizontal bottom part of the TSD eases its ``insertion’’ between the two front wheels of the wheelchair}.

\section{Conclusion}

This article presented the humanoid robot RHP Friends, detailing the hardware and software components that enable its autonomous locomanipulation and perception of large objects, multi-contact planning and control, and immersive teleoperation. These cutting-edge technologies were integrated together to demonstrate a combination of autonomous robotics and robot teleoperation within a nursing care context where some tasks, such as transferring a person from a bed to a wheelchair, can be automated. In contrast, unpredictable tasks, such as partial power outages, can be handled through teleoperation. Switching between the two modes can be done in seconds. This combination presents a unique challenge because the humanoid platform must allow the locomanipulation of a loaded device weighing about the same as the robot and achieve fine manipulation of a switch thinner than the tip of the robot's gripper. Performing the demonstration several times a day, in front of a large audience provided also a valuable opportunity to validate the robustness and maturity of the system components under real-world conditions while also highlighting areas that need significant improvement, especially in ensuring dependable performance under such high-pressure situations, for example not being able to choose when to start the experiment, or having only the robot sensors as feedback.

With regard to the complexity of the system, while our current teleoperation system demonstrates the feasibility of using consumer-grade equipment for remote intervention in healthcare, we recognize that it still has significant limitations. Moving forward, our vision is to develop a more seamless and intuitive interface that can adapt to diverse healthcare scenarios, offering both ease of use and the fine control required for delicate tasks.

Increasing the speed of execution is also crucial for practical use. While ensuring patient safety remains a primary concern, particularly during close contact, there are phases where higher speed can be achieved, such as when the robot is navigating or performing tasks without direct patient interaction. To address this, we plan to explore shared control approaches that leverage a fast visual tracker to assist the teleoperator, improving precision and reducing execution time. Additionally, we may enhance the efficiency by  improving the hardware and integrating a whole-body Model Predictive Control (MPC) framework to better manage induced inertia and dynamic responses, ultimately making the robot's actions more fluid and efficient.

\bibliographystyle{IEEEtran}
\bibliography{bare_jrnl.bib}

\end{document}